\documentclass[conference]{IEEEtran}
\usepackage{graphicx}
\usepackage{threeparttable}
\usepackage{cite}
\ifCLASSINFOpdf
\else
\fi
\begin{document}
\title{InterpolationSLAM: A Novel Robust Visual SLAM System in Rotating Scenes}

\author{\IEEEauthorblockN{Zhenkun Zhu}
\IEEEauthorblockA{Department of Automation\\
University of Science and Technology of China\\
Hefei, China 230027\\
Email: zhuzhenkun@mail.ustc.edu.cn}
\and
\IEEEauthorblockN{Jikai Wang}
\IEEEauthorblockA{Department of Automation\\
University of Science and Technology of China\\
Hefei, China 230027\\
Email: wangjk@ustc.edu.cn}
\and
\IEEEauthorblockN{Zonghai Chen}
\IEEEauthorblockA{Department of Automation\\
University of Science and Technology of China\\
Hefei, China 230027\\
Telephone: +86-0551-63606104\\
Email: chenzh@ustc.edu.cn}}


%Telephone: (800) 555--1212, Fax: (888) 555--1212}


\maketitle

\begin{abstract}
In recent years, visual SLAM has achieved great progress and development, but in complex scenes, especially rotating scenes, the error of mapping will increase significantly, and the slam system is easy to lose track. In this article, we propose an InterpolationSLAM framework, which is a visual SLAM framework based on ORB-SLAM2. InterpolationSLAM is robust in rotating scenes for Monocular and RGB-D configurations. By detecting the rotation and performing interpolation processing at the rotated position, pose of the system can be estimated more accurately at the rotated position, thereby improving the accuracy and robustness of the SLAM system in the rotating scenes. To the best of our knowledge, it is the
first work combining the interpolation network into a Visual SLAM system to improve SLAM system robustness in rotating scenes. We conduct experiments both on KITTI Monocular and TUM RGB-D datasets. The results demonstrate that InterpolationSLAM outperforms the accuracy of standard Visual SLAM baselines.
\end{abstract}


% \fi
\IEEEpeerreviewmaketitle

\section{Introduction}
Simultaneous localization and mapping(SLAM) has become a research hotspot in recent years and plays an important role in different fields, such as autonomous driving, AR, VR and so on. According to different sensors, SLAM can be divided into visual SLAM(VSLAM) and lidar SLAM. VSLAM has its own advantage for its wide applicability and low cost. Among them, monocular SLAM only relies on one camera to complete all the functions. Since the depth of objects cannot be obtained through binocular vision or depth cameras, its accuracy and robustness are not high, especially in rotating scenes, pose estimation and accuracy of mapping will decrease significantly.

Traditional VSLAM relies heavily on the quality and quantity of pictures. In rotating scenes, the turning angle between adjacent frames is too large, resulting in a large field of view difference, and the number of matched feature points will drastically reduce, which is likely to cause tracking loss or drift. The success of deep learning methods in image processing field has led people to use deep learning methods to improve the performance of SLAM systems. In response to improve the robustness of feature points, researchers began to use neural networks to train more robust feature points in order to improve the performance of SLAM system. SuperPoint-VO\cite{han2020superpointvo} uses SuperPoint\cite{detone2018superpoint} to extract features and integrates them into a visual odometer. SuperPoint uses Sythesis Dataset and Homographic Adaptation strategy training to extract feature points, which have good performance for indoor scenes, but it is easy to get lost or mismatches for outdoor objects, especially non-geometric objects.
DF-SLAM\cite{kang2019df} preserves the FAST\cite{rosten2006machine} keypoints and uses CNN to extract descriptors. DXSLAM\cite{li2020dxslam} uses HF-Net\cite{sarlin2019coarse} to extract features from each image frame. HF-Net is able to predict keypoint detection, dense local descriptors, and global descriptors respectively. The fusion of local features and global features makes DXSLAM more robust under environmental changes and viewing angle changes.

The generalization of features extracted by deep learning methods has always been a potential problem. The traditional hand-crafted feature point has good generalization ability, but its performance will decrease significantly in rotating scenes. In response to this problem, we propose a new type of SLAM system called InterpolationSLAM. By conducting interpolation strategy in the rotating scenes, the field of view transformation between adjacent frames is reduced, so that the feature matching is more stable, thereby improving the robustness and accuracy of the visual SLAM system. In summary, our contribution can be divided into the following three aspects:

1. As far as we know, we are the first to apply the interpolation neural network to the SLAM system to improve the performance of the SLAM system in rotating scenes.

2. Through the detection function, we only perform interpolation processing during the rotation process and only in the fast rotating scenes.

3. Our system has been tested on the famous KITTI\cite{geiger2012we} and TUM datasets\cite{sturm2012benchmark}, and achieved a good result.

\section{Related Work}
The purpose of our work is to improve the performance of SLAM system in rotating scenes with frame interpolation technology. This method mainly involves two parts, including frame interpolation and VSLAM.

\subsubsection{Frame interpolation}
Frame interpolation\cite{parihar2021comprehensive} is mainly served for video post-processing, surveillance, and video restoration tasks. It aims to increasing the frame rate of a video sequence by calculating intermittent frames between consecutive input frames. Advanced deep learning algorithms have the potential to discover knowledge from large-scale diverse video data. Long et al.\cite{long2016learning} first apply CNN to do optical flow estimation and predict the intermediate frames. They design an auto-encoded network structure similar to Flownet-S\cite{dosovitskiy2015flownet}. Niklaus et al.\cite{niklaus2017video} realize pixel-level frame interpolation and they calculate a separate kernel for each pixel. Besides, they use convolution kernels combine motion estimation and re-sampling to one step, making the proposed neural network end-to-end trainable. Vidanpathirana et al.\cite{vidanpathirana2020tracking}attempt to reduce optical flow errors by designing a pose tracking system. They also provided a fast point tracking solution to accelerate the system. However, calculating 2D convolution kernels for each pixel costs too much memory and time, which does not satisfy the requirements of high-resolution videos. Niklaus et al.\cite{niklaus2017video2} improve on \cite{niklaus2017video} and replace 2D convolution kernels with 1D convolution kernels for each pixel. They also design a dedicated encoder–decoder neural network to estimate kernels for all pixels in a frame at one time. Xue et al.\cite{xue2016visual} introduces cross-convolution layer to guide network to learn from feature maps and kernel weights, and the proposed network is able to predict multiple extrapolated frames from a single frame. Liu et al.\cite{liu2017video} uses dense voxel flow, which is similar to optical flow method but it also considers time component, for frame interpolation. Liu et al.\cite{liu2019deep} creates cycle consistency loss to train interpolation network and the novel function can better preserve the motion information. Lee et al\cite{lee2020adacof} designed a new warping module which only performs interpolation on target pixels and locations to decrease interpolation time.

\subsubsection{VSLAM}
VSLAM can be mainly divided into feature-based method and direct method. The direct method is based on the assumption of gray scale invariant, and has insufficient adaptability to complex environments. The feature-based method is based on feature extraction and matching for pose estimation and mapping, which has better robustness. The method proposed in this paper is developed based on the feature-based method.

In recent years, combining neural network to improve SLAM system's performance has become a hotspot. Other than using neural network to extract high-quality feature, Some scholars try to use semantic information to improve SLAM system. One part of them are aimed to improving the positioning accuracy and robustness of the SLAM system based on semantic information. Konstantinos-Nektarios et al.\cite{lianos2018vso} propose VSO, which applys scene semantic information to establish mid-term constraints in the tracking process, thereby reducing the scale drift of the visual odometer and improving the positioning accuracy. Other part of these scholars use semantic information to build semantic maps with richer information. Margarita et al.\cite{grinvald2019volumetric} propose to project the structure of Mask RCNN\cite{he2017mask} to a 3D perspective for instance inference and thus build the map with more details. What's more, Kendall et al.\cite{kendall2015posenet}design an end-to-end deep learning model PoseNet to solve the task of camera pose prediction. PoseNet model introduces the idea of absolute pose regression using neural networks. It uses GoogLeNet\cite{szegedy2015going} network as a scene expression model, calculating the weighted sum of position error and attitude error according to predefined weights, and uses it as a pose loss function to supervise position prediction and attitude prediction. It is an important trend to combine VSLAM with neural network in the future.

\section{System Description}
In this part, we will introduce InterpolationSLAM in detail. This part mainly includes four aspects. First, a block diagram of the entire InterpolationSLAM system structure will be given. Then, we will introduce how to identify rotating scenes and whether the rotation is fast enough to perform interpolation. Then we briefly introduce the process of InterpolationSLAM using neural networks for interpolation. Finally, we introduce the entire positioning and tracking process.

\subsubsection{Framework of InterpolationSLAM}

As an outstanding representative of feature-based method SLAM, ORB-SLAM2\cite{mur2017orb} has achieved good results on both outdoor and indoor datasets. It provides three modes: monocular mode, stereo mode and RGB-D mode. Our system is based on ORB-SLAM2\cite{mur2017orb}. Fig. 1 shows the architecture of our entire SLAM system. RGB images in monocular or RGB-D modes are first input into the SLAM system, and then feature extraction and matching are performed to estimate the pose. The rotation detection is used to detect whether the system is rotating and whether the rotating speed exceeds the threshold. If it exceeds the threshold, interpolation network begins to interpolate between the current image and the next image, we do not perform rotation detection or interpolation processing on the predicted intermediate image, when the rotation speed is less than the threshold, the interpolation process ends.

\begin{figure*}
\centering
\includegraphics[width=7.5in]{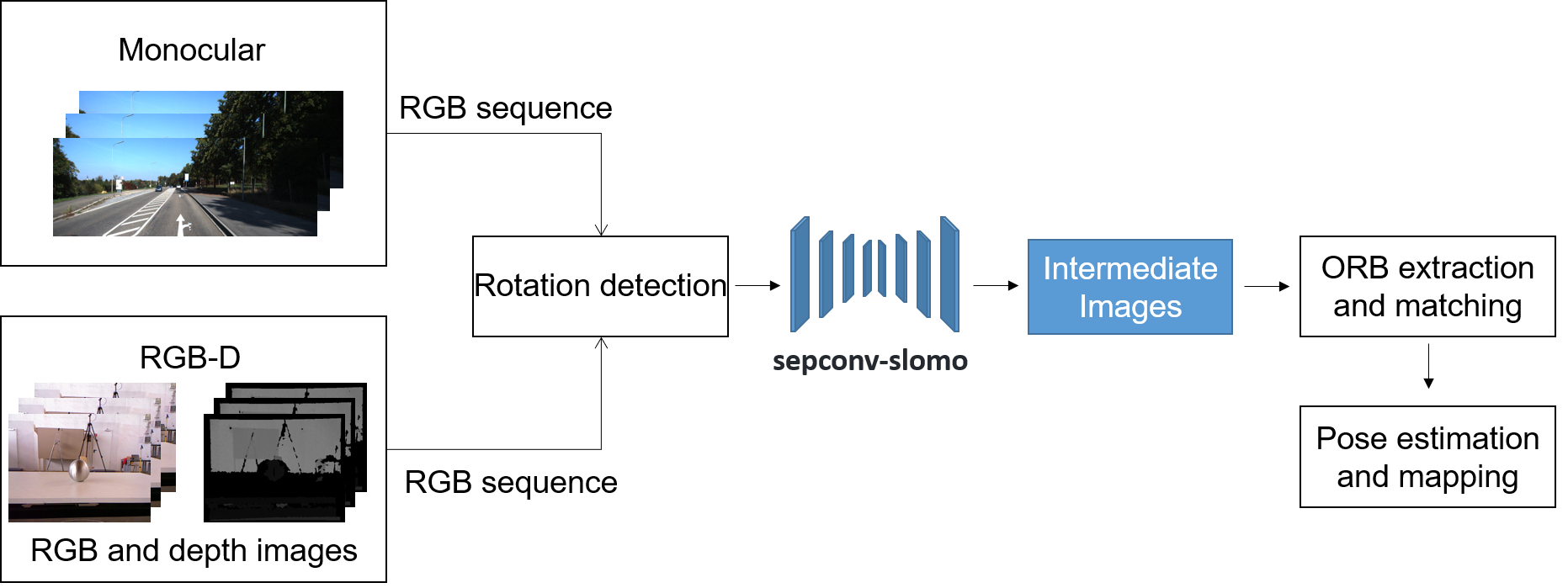}
\caption{Framework of InterpolationSLAM.}
\label{fig_kitti02_traj}
\end{figure*}

\subsubsection{Rotation Detection}

The PoseOptimization function in ORB-SLAM2 is used to obtain the transformations $T_{{cw}_{t-1}}$ and $T_{{cw}_t}$ from the world coordinate system to the camera coordinate system for previous frame $I_{t-1}$ and the current frame $I_t$ respectively, and then the pose transformation from the camera coordinate system of the previous frame to the camera coordinate system of the current frame $T_{c_t c_{t-1}}$ can be obtained:
\begin{equation}
T_{c_t c_{t-1}}=T_{{cw}_t}*T_{{cw}_{t-1}}^{-1}
\end{equation}
where $T_{{cw}_{t-1}}^{-1}$ means the inverse of $T_{{cw}_{t-1}}$and then the 3*3 matrix in the upper left corner of the $T_{c_t c_{t-1}}$ matrix is the rotation matrix $R_{c_t c_{t-1}}$. The corresponding Euler angle $\theta$ of $R_{c_t c_{t-1}}$ can be further solved by:
\begin{equation}
\theta=arccos\frac{tr(R_{c_t c_{t-1}})-1}{2}
\end{equation}
When $\theta>\beta$, we think that it encounters a turn and the turning speed is relatively fast, so the interpolation process starts. In our experiments, for KITTI datasets, we set $\beta$ equal to 0.03.

\subsubsection{Interpolation}
In order to interpolate images, we use the Adaptive Convolution framework, which greatly improves the interpolation speed by integrating motion estimation and pixel synthesis into one step. Through testing and analysis, we use sepconv-slomo network\cite{niklaus2017video2} for interpolation processing. The processing speed and processing effect show good performance.

The input of sepconv-slomo is RGB raw image. Our idea is to use interpolation to reduce the field of view difference in rotating scenes so as to improve the performance of SLAM system in rotating scenes.

\subsubsection{Tracking}
Images generated by interpolation are the same format as the original images. They are fed into the SLAM system for feature extraction, matching and so on to complete the whole process of VSLAM.

\section{Experimental Results}
In this part, we demonstrate the performance of the InterpolationSLAM system through experiments. The effectiveness of the system is proved through the trajectory and EVO evaluation results, and it is quantitatively confirmed that the proposed method can improve the number of feature matching pairs and matching quality. By comparing with ORB-SLAM2, we found that a significant performance improvement has been achieved in rotating scenes.

\subsubsection{Evaluation Using KITTI Datasets}
KITTI datasets are currently recognized large-scale outdoor scene datasets, which provide data for monocular SLAM, stereo SLAM, target detection, semantic segmentation and so on. KITTI contains multiple scenes such as cities, villages, and highways. We perform interpolation processing on the KITTI dataset using monocular mode. Due to the time-cost consideration of the SLAM system and the small error of ORB-SLAM2 when going straightforward, we perform rotation detection on it. Only when the rotating speed exceeds threshold 0.03rad/s, the interpolation strategy is enabled. KITTI04 sequence does not have obvious rotating scenes. The test results of other sequences are shown in the Table 1. The results show that our SLAM system has improvements in 6 sequences, and achieved significant improvementa in the KITTI02 and KITTI10 sequence. The trajectory of KITTI02 is shown in Fig. 2. It can be seen from the trajectory that, compared to ORB-SLAM2, InterpolationSLAM has a significant improvement in multiple rotating scenes(turning spots). We use the trajectory and error on KITTI10 for further explanation. The trajectory of ORB-SLAM2 has a large deviation from the goundtruth at the red circle on the left in Fig. 3, while the trajectory of InterpolationSLAM here is almost the same as the ground truth. The red circle on the right side of Fig. 3 corresponds to the error caused in this turning spot.

\begin{table}
\caption{Evaluation On KITTI}
\label{table_example}
\centering
% Some packages, such as MDW tools, offer better commands for making tables
% than the plain LaTeX2e tabular which is used here.
\begin{tabular}{|c||c||c|}
\hline
sequence & ORB-SLAM2 & InterpolationSLAM\\
\hline
00 & 0.307741 & 0.354470\\
\hline
01 & 2.059531 & 2.861581\\
\hline
02 & 1.469590 & 0.409261\\
\hline
03 & 0.100683 & 0.115383\\
\hline
05 & 0.261811 & 0.238920\\
\hline
06 & 0.640808 & 0.583717\\
\hline
07 & 0.241249 & 0.226254\\
\hline
08 & 5.193368 & 6.244322\\
\hline
09 & 4.020554 & 3.616012\\
\hline
10 & 0.523648 & 0.224900\\
\hline
\end{tabular}
\end{table}

\begin{figure*}
\centering
\includegraphics[width=7.5in]{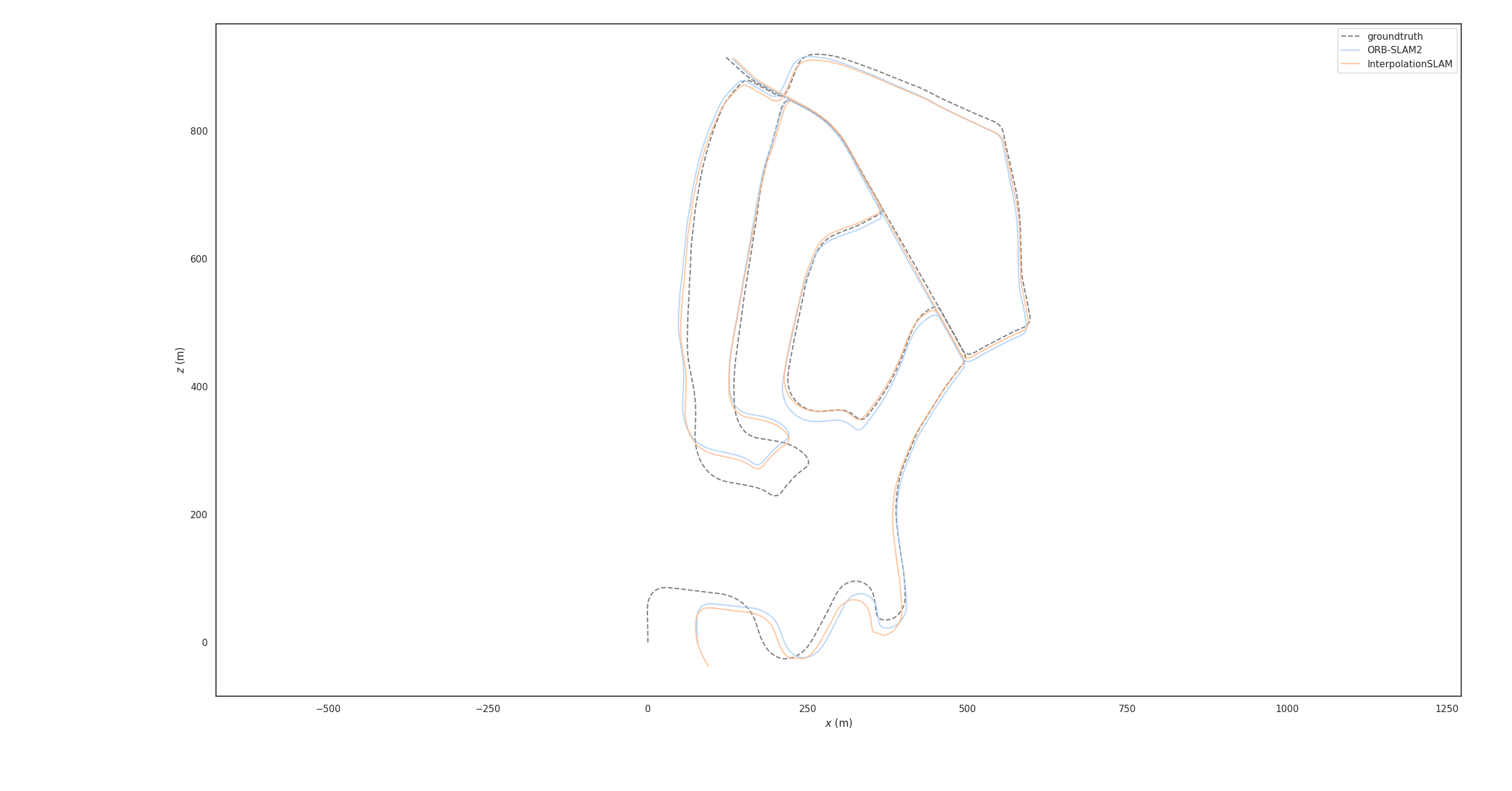}
\caption{Trajectory of ORB-SLAM2 and InterpolationSLAM in KITTI02.}
\label{fig_kitti02_traj}
\end{figure*}

\begin{figure*}
\centering
\includegraphics[width=7.5in]{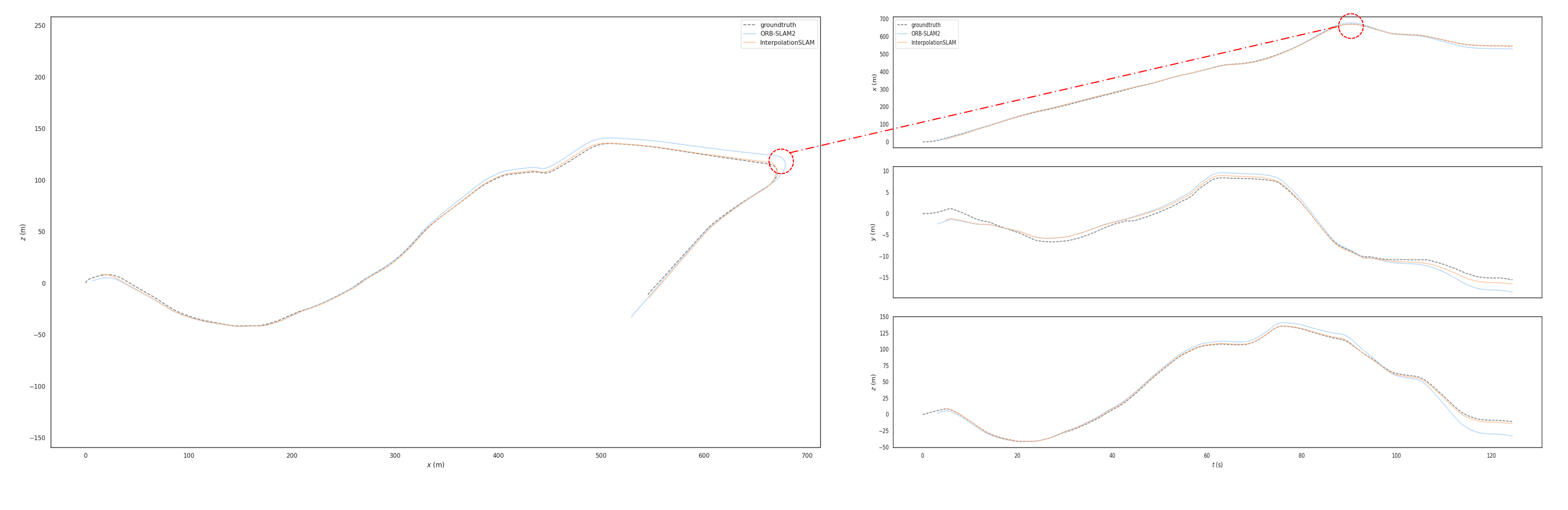}
\caption{Trajectory and error of ORB-SLAM2 and InterpolationSLAM in KITTI10.}
\label{fig_kitti10_traj}
\end{figure*}

\subsubsection{Evaluation Using TUM RGB-D Datasets}
The TUM datasets are currently recognized indoor scene datasets, which provide RGB images and depth images. We test in the sequence with many rotations, using interpolation in the whole sequence, and the result is shown in the Table 2.

\begin{table}
\caption{Evaluation On TUM RGB-D}
\label{table_example}
\centering
% Some packages, such as MDW tools, offer better commands for making tables
% than the plain LaTeX2e tabular which is used here.
\begin{threeparttable}
\begin{tabular}{|c||c||c|}
\hline
sequence & ORB-SLAM2 & InterpolationSLAM\\
\hline
fr1\_360 & 0.160174 & 0.136622\\
\hline
fr2\_coke & 0.650956 & 0.343787\\
\hline
fr2\_large\_no\_loop & 0.148426 & 0.09812\\
\hline
fr2\_metallic\_sphere & 0.523895 & 0.586258\\
\hline
fr2\_metallic\_sphere2 & 0.149060 & 0.052703\\
\hline
fr2\_pioneer\_360 & - & 0.020144\\
\hline
fr3\_nostructure\_texture\_far & 0.063058 & 0.040973\\
\hline
fr3\_teddy & 0.021636 & 0.015174\\
\hline
\end{tabular}
\footnotesize
- means losing track
\end{threeparttable}
\end{table}

In the fr2\_pioneer\_360, while ORB-SLAM2 lose track, through interpolation processing, InterpolationSLAM can complete positioning and mapping on the entire sequence. So as to further analyze the reason, we analyze the image feature matching before and after interpolation as shown in Fig. 4. The above is the original matching result, the number of match pairs is 43, and it can be intuitively seen that the matching quality is not good. The following is the matching effect after interpolation. There are a total of 94 matching pairs between last frame and an intermediate frame, and a total of 53 matching pairs in the intermediate frame and the current frame, and the improvement of the matching quality can be clearly seen.

\begin{figure*}[t]
\centering
\includegraphics[width=7.5in]{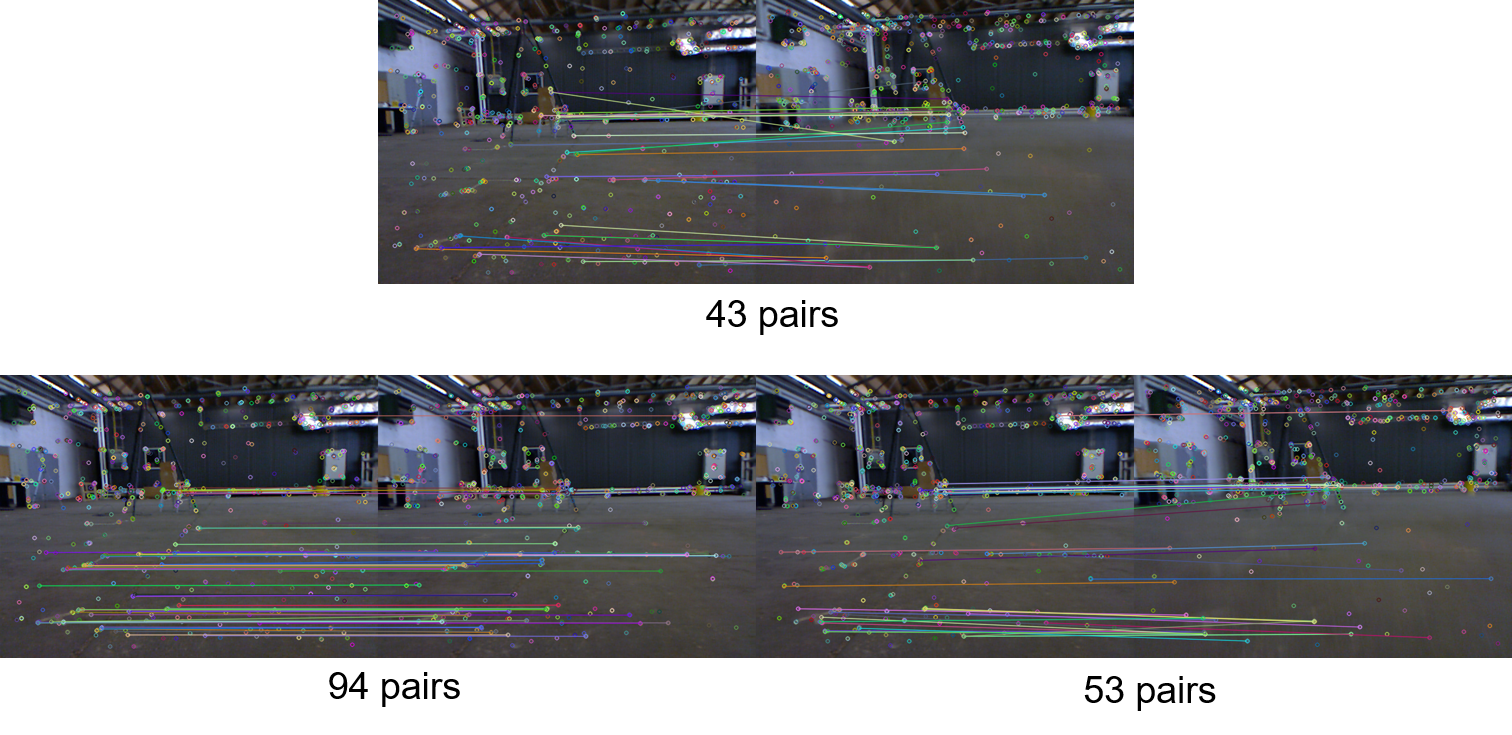}
\caption{Interpolation effect in fr2\_pioneer\_360.}
\label{fig_kitti10_traj}
\end{figure*}

The trajectory and error of ORB-SLAM2 and InterpolationSLAM in fr2\_metallic\_sphere2 and fr2\_pioneer\_360 are shown in Fig. 5 and Fig. 6 respectively.
\begin{figure*}[t]
\centering
\includegraphics[width=7.5in]{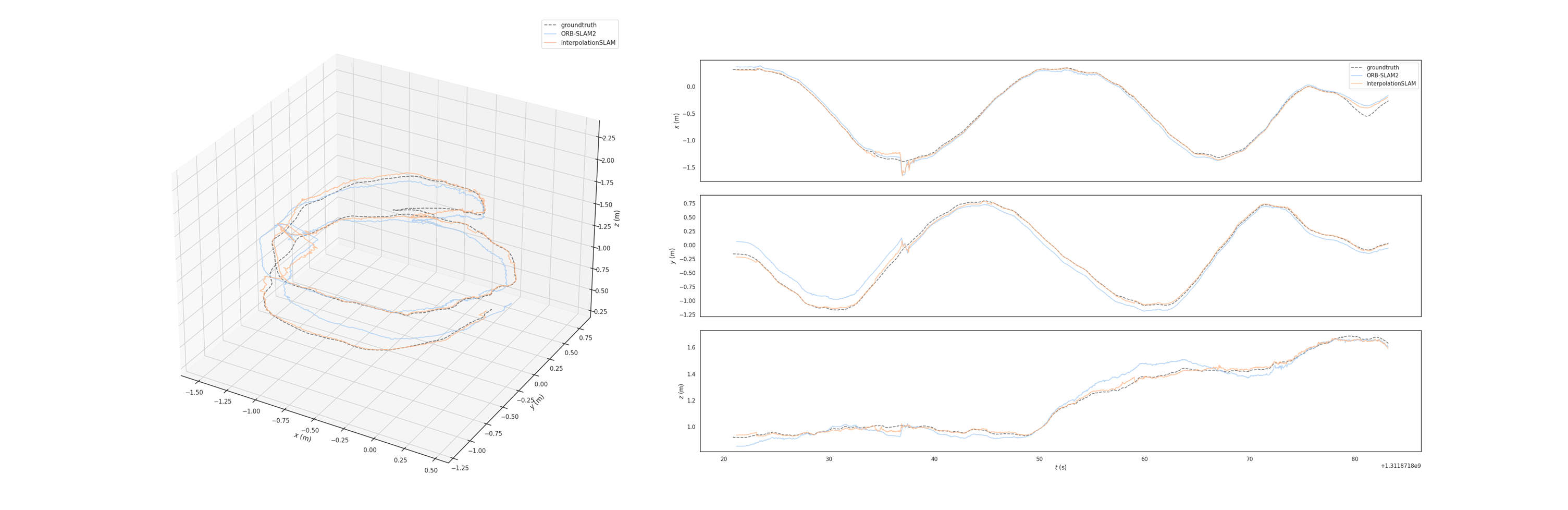}
\caption{Trajectory and error of ORB-SLAM2 and InterpolationSLAM in fr2\_metallic\_sphere2.}
\label{fig_kitti10_traj}
\end{figure*}

\begin{figure*}[t]
\centering
\includegraphics[width=7.5in]{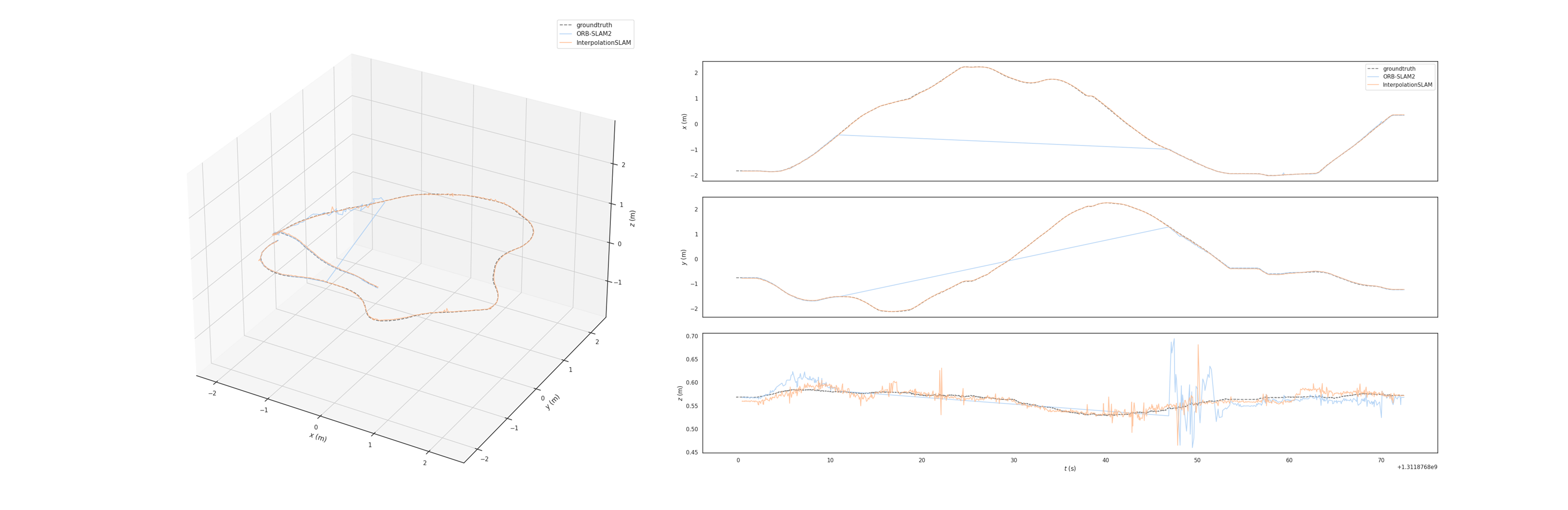}
\caption{Trajectory and error of ORB-SLAM2 and InterpolationSLAM in fr2\_pioneer\_360.}
\label{fig_kitti10_traj}
\end{figure*}

\section{Conclusion}
we have presented a Visual SLAM system based on ORB-SLAM2, which adds a frame interpolation network, making the SLAM system more robust in rotating scenes for monocular and RGB-D cameras. Through frame interpolation, the field of view change between adjacent frames is significantly narrowed and the pose estimation is more accurate in the later period. Comparing with ORB-SLAM2, our method shows better performance.

Rotation detection seems a bit rigid, and adaptive strategies can be further adopted. At the same time, this article only studies the effect of interpolation on turning, and interpolation may also further improve SLAM performance in lighting change scenes. How to improve the speed of interpolation is also a good research direction.


\section*{Acknowledgment}

The authors would like to thank...



%\bibliographystyle{IEEEtran}
\bibliographystyle{IEEEtran}
\bibliography{reference.bib}


% \end{thebibliography}

\end{document}